\def\PREPRINT{1} 

\ifnum\PREPRINT=1
    \documentclass{llncs}
    \usepackage{times,fullpage}
\else
    \documentclass[pdflatex,sn-mathphys]{sn-jnl}
\fi

\usepackage{graphicx}
\usepackage{color}
\usepackage{algorithm}
\usepackage{algpseudocode}
\usepackage{appendix}
\usepackage{booktabs}
\usepackage{array}
\usepackage{float}
\usepackage{array}
\usepackage{enumitem}
\usepackage{footmisc}
\usepackage{amsmath}
\usepackage{multirow}
\usepackage{longtable}
\usepackage{comment}
\usepackage[T1]{fontenc}
\usepackage{url}
\ifnum\PREPRINT=1
    \usepackage[square,numbers]{natbib}
    \usepackage{parskip} 
\else
    \usepackage{natbib}
\fi

\ifnum\PREPRINT=0
    \jyear{2022}%
    
    \theoremstyle{thmstyleone}%
    \theoremstyle{thmstyletwo}%
    \theoremstyle{thmstylethree}%
\fi

\begin{document}

\ifnum\PREPRINT=0
    \title[Coevolution and substitution of the fittest for recommender systems]{Using coevolution and substitution of the fittest for health and well-being recommender systems}
    
    \author*[1]{\fnm{Hugo} \sur{Alcaraz-Herrera}}\email{h.alcarazherrera@bristol.ac.uk}
    \author[1]{\fnm{John} \sur{Cartlidge}}\email{john.cartlidge@bristol.ac.uk}
    \equalcont{These authors contributed equally to this work.}
    \affil*[1]{\orgdiv{Department of Computer Science}, \orgname{University of Bristol}, \orgaddress{ \country{UK}}}
\else

    \title{Using coevolution and substitution of the fittest for health and well-being recommender systems}
    \author{
    Hugo Alcaraz-Herrera\inst{1}
    \and 
    John Cartlidge\inst{1}}
    \institute{Department of Computer Science, University of Bristol, Bristol, UK. \\
    \email{h.alcarazherrera@bristol.ac.uk,john.cartlidge@bristol.ac.uk}}
    
    \maketitle
\fi

\abstract{This research explores substitution of the fittest (SF), a technique designed to counteract the problem of disengagement in two-population competitive coevolutionary genetic algorithms. SF is domain-independent and requires no calibration. We first perform a controlled comparative evaluation of SF's ability to maintain engagement and discover optimal solutions in a minimal toy domain. Experimental results demonstrate that SF performs similarly to alternative techniques presented in the literature but has the advantage of requiring no parameter tuning. 
We then address the more complex real-world problem of evolving recommendations for health and well-being. We introduce a coevolutionary extension of EvoRecSys, a previously published evolutionary recommender system. We demonstrate that SF is able to maintain a better trade-off between engagement and performance than other techniques in the literature, and the resultant recommendations using SF are higher quality and more diverse than those produced by EvoRecSys. 
}




\ifnum\PREPRINT=0
    \maketitle
\else
    \setlength{\parindent}{20pt} 
\fi

\section{Introduction}\label{sec:introduction}

While attempting the problem of designing optimal sorting networks using a genetic algorithm (GA), Hillis decided that rather than use randomly generated input lists to evaluate sorting networks, he would instead co-evolve a population of input lists that are evaluated on their ability to not be sorted \cite{Hillis1990}. By coupling the evolution of input lists with the evolution of the networks to sort those lists, Hillis attempted to create an ``arms race'' dynamic such that input lists consistently challenge networks that are sorting them. As networks improve their ability to sort, so lists become more difficult to sort, etc. This {\em coevolutionary} approach significantly improved results and generated wide interest, offering a new ability to tackle domains where an evaluation function is unknown or difficult to operationally define; and offering the potential for open-ended evolutionary progress through self-learning. 

However, it soon emerged that coevolution can suffer from  pathologies that cause the system to behave in an unwanted manner and prevent continual progress towards some desired goal. 
For example, populations may continually cycle with no overall progress; populations may progress in an unintended and unwanted direction; populations may stop competing and settle into a mediocre stable state; and populations may disengage and stop progressing entirely. These pathologies have been studied in depth and a variety of techniques have been introduced as remedy (see \cite{Popovici2012}). Yet, there is still much to be understood, and no panacea has been discovered. 

In previous work \cite{AlcarazH2021}, the current authors proposed {\em substitution of the fittest} (SF), a novel domain-independent method designed to tackle the problem of disengagement in two-population competitive coevolutionary systems. In \cite{AlcarazH2021}, SF was explored in the deliberately simple {\em greater than} domain, a toy domain originally introduced to elaborate the dynamics of coevolution \cite{Watson2001}. In this paper, we extend \cite{AlcarazH2021} in two ways. First, we perform a more extensive comparative analysis of SF in the greater than domain. Second, and more significantly, we address the more complex problem of recommender systems for health and well-being, and demonstrate that a coevolutionary approach using SF outperforms the evolutionary approach published in \cite{AlcarazH2022}. 

The rest of this paper is organised as follows. Section~\ref{sec:background} presents a review of coevolutionary genetic algorithms, their pathologies, and techniques to counteract said pathologies.  Much of this background is reproduced directly from \cite{AlcarazH2021}. In Section~\ref{sec:sf}, we introduce substitution of the fittest. Section~\ref{sec:co} presents a series of controlled experiments conducted in the deliberately minimal ``greater than'' domain and the performance of SF is compared against other techniques from the literature, {\em reduced virulence} (RV) \cite{Cartlidge2004b} and {\em autonomous virulence adaptation} (AVA) \cite{Cartlidge2011}. In Section~\ref{sec:cers}, we present experiments conducted in the more complex real-world domain of recommender systems for health and well-being. This work introduces a coevolutionary extension of a  previously published evolutionary recommender system (EvoRecSys) \cite{AlcarazH2022}. It is shown that SF is able to produce recommendations that are of higher quality and more diverse than the results obtained using RV and AVA, and also produces better overall recommendations than the standard evolutionary approach of EvoRecSys. Finally, Section~\ref{sec:conclusions} presents conclusions and avenues for future research investigation.    

\section{Background}\label{sec:background}

Coevolutionary genetic algorithms with two distinct populations are often described using terminology that follows the biological literature. 
As such, and following Hillis' original formulation, the populations are often named as ``hosts'' and ``parasites'' \cite{Hillis1990}. 
In such cases, the host population tends to denote the population of candidate ``solutions'' that we are interested in optimising (e.g., the sorting networks), while the parasite population tends to denote the population of test ``problems'' for the solution population to solve (e.g., the lists to sort); i.e., the hosts are the {\em models} and the parasites are the {\em training set}; or alternatively the hosts are the {\em learners} and the parasites are the {\em teachers}. 
Throughout this paper, we tend to use the host-parasite terminology to distinguish coevolving populations. 
However, while this terminology is meaningful in {\em asymmetric} systems where one population (the model) is of most interest, it should be noted that in {\em symmetric} systems, such as games of self-play where both coevolving populations are models with the same encoding scheme, the two populations become interchangeable and the names {\em host} and {\em parasite} have less meaning. In an ideal scenario, two-population competitive coevolution will result in an {\em arms race} such that both populations continually evolve beneficial adaptions capable of outperforming competitors. 

As a result, there is continual system progress towards some desired optimum goal. However, this ideal scenario rarely materialises. In practice, coevolutionary systems tend to exhibit pathologies that restrict progress \cite{Watson2001}. These include {\em cycling}, where populations evolve through repeated trajectories like players in an endless game of {\em rock-paper-scissors}; and while short-term evolution exhibits continual progress, there is no long-term global progress \cite{Cartlidge2004a}.

Alternatively, populations may start to {\em overspecialise} on sub-dimensions of the game, such that evolved solutions are brittle and do not generalise \cite{Cartlidge2003}. Furthermore, one population may begin to dominate the other to such an extent that populations {\em disengage} and evolutionary progress fails altogether, with populations left to drift aimlessly \cite{Cartlidge2004b}. The likelihood of suffering from these pathologies can be exacerbated by the problem domain. Cycling is more likely when the problem exhibits {\em intransitivity}; overspecialisation is more likely in {\em multi-objective} problems; and disengagement is more likely if the problem has an asymmetric {\em bias} that favours one population \cite{Watson2001}. 

Numerous techniques have been proposed for mitigating the pathologies that prevent continual coevolutionary progress (for detailed reviews, see  \cite{Popovici2012,Miguel2018}). We can roughly group these approaches into three broad categories; although in practice many techniques straddle more than one category.

First, there are {\em archive} methods, which are designed to preserve potentially valuable adaptations from being ``lost'' during the evolutionary process. The first coevolutionary archiving technique is the {\em Hall of Fame} (HoF) \cite{Rosin1997b}. Every generation, the elite member of each population is stored in the HoF archive. Then, individuals in the current population are evaluated against current competitors and also against members of the HoF. This ensures that later generations are evaluated on their capacity to beat earlier generations as well as their contemporaries. However, as the archive grows each generation, simple archiving methods like the HoF can become unwieldy over time. To counter this, more sophisticated and efficient archiving methods have been introduced to simultaneously minimise archive size while maximising  archive ``usefulness''. An efficient example is the Layered Pareto Coevolutionary Archive (LAPCA), which only stores individuals that are {\em non-dominated} and {\em unique}; while the archive itself is pruned over time to keep the size within manageable bounds \cite{DeJong2007}. 

More recent variations on Pareto archiving approaches include rIPCA, which has been applied to the problem of network security through the coevolution of adversarial network attack and defence dynamics \cite{Garcia2017}. Pareto dominance has also been  employed for selection without the use of an archive, for example the Population-based Pareto Hill Climber \cite{Bari2018}; and Pareto fronts have been incorporated into an ``extended elitism'' framework, where offspring are selected only if they Pareto dominate parents when evaluated against the {\em same} opponents \cite{Akinola2020}. 

A second popular class of approaches attempt to maintain a diverse set of evolutionary challenges through the use of {\em spatial embedding} and {\em multiple populations}. Spatially embedded algorithms -- where populations exist on an n-dimensional plane and individuals only interact with other individuals in the local neighbourhood -- have been shown to succeed where other non-spatial coevolutionary approaches fail. Explanations for how spatial models can help combat disengagement through challenge diversity have been explored in several works \cite{Wiegand2004,WilliamsC2005}.
Challenge diversity can also be maintained through the use of multiple genetically-distinct populations (i.e., with no interbreeding or migration). Examples include the \emph{friendly competitor}, where two {\em model} populations (one ``friendly'' and one ``hostile'') are coevolved against one {\em test} population \cite{Ficici98a}. Tests are rewarded if they are both easy to be defeated by a friendly model and hard to be beaten by a hostile model; thereby ensuring pressure on tests to evolve at a challenge-level consistent with the ability of models. Recently, a new method incorporating the periodic spawning of sub-populations, and then re-integration of individuals that perform well across multiple sub-populations back into the main population has been shown to encourage continual progress in predator-prey robot coevolution \cite{Simione2021}.

Finally, there are approaches that focus on adapting the {\em mechanism for selection} such that individuals are not selected in direct proportion to the number of competitions that they win; i.e., selection favours individuals that are {\em not} unbeatable. An early endeavour in this area is the {\em phantom parasite}, which marginally reduces the fitness of an unbeatable competitor, while all other fitness values remain unchanged \cite{Rosin1997}. Later, the $\Phi$ function was introduced for the density classification task to coevolve cellular automata rules that classify the density of an initial condition \cite{Pagie2002}. The $\Phi$ function translates all fitness values such that individuals are rewarded most highly for being equally difficult and easy to classify (i.e., by being classified correctly half of the time); while individuals that are always classified or always unclassified are punished with low fitness. However, while $\Phi$ worked well, it was limited by being domain-specific. 

More generally applicable is the {\em reduced virulence} (RV) technique \cite{Cartlidge2002,Cartlidge2004b}. Inspired by the behaviour of biological host-parasite systems, where the virulence of pathogens evolves over time, reduced virulence is the first domain-independent technique with tunable parameters that can be configured. After generating a parasite score through competition, reduced virulence applies the following non-linear function to generate a fitness for selection:

\begin{equation}\label{eq:rv}
    f(x_i,\upsilon) = \frac{2x_i}{\upsilon} - \frac{x_i^2}{\upsilon^2}
\end{equation}
\vspace{0.1cm}

\noindent
where $0 \leq x_i \leq 1$ is the relative (or subjective) fitness of individual $i$ and $0.5 \leq \upsilon \leq 1$ represents the virulence of the parasite population. When $\upsilon=1$, equation~(\ref{eq:rv}) preserves the original ranking of parasites (i.e., the ranking of competitive score, $x$) and is equivalent to the canonical method of rewarding parasites for all victories over hosts. 
When $\upsilon=0.5$, equation~(\ref{eq:rv}) rewards maximum fitness to parasites that win exactly half of all competitions. 
Therefore, in domains where there is a bias in favour of one population (the ``parasites''), setting a value of $\upsilon<1$ for the advantaged population reduces the bias differential in order to preserve coevolutionary engagement. 
Reduced virulence demonstrated improved performance, but is limited by requiring $\upsilon$ to be determined in advance. 
In many domains, bias may be difficult to determine and may change over time. 
To tackle this problem, reduced virulence has been incorporated into a human-in-the-loop system enabling a human controller to {\em steer} coevolution during runtime by observing the system behaviour and altering the value of $\upsilon$ in real time \cite{Bullock2002}.

Later, {\em autonomous virulence adaptation} (AVA) -- a machine learning approach that automatically updates virulence during coevolution -- was proposed \cite{Cartlidge2011}. Each generation $t$, AVA updates $\upsilon$ using:

\begin{equation}\label{eq:ava_v_t_plus_one}
    \upsilon_{t+1} = \upsilon_{t} + \Delta_t
\end{equation}

\begin{equation}\label{eq:ava_delta_t}
    \Delta_t = \mu\Delta_{t-1} + \alpha(1 - \mu)(\tau - \overline{X_t})
\end{equation}
\vspace{0.1cm}

\noindent 
where $0 \leq \alpha, \mu, \tau \leq 1$ are learning rate, momentum, and target value, respectively; and $\overline{X_t}$ is the normalised mean subjective score of the population.\footnote{To avoid immediate disengagement in cases of extreme bias differential, for the initial $t<5$ generations   equation~(\ref{eq:ava_delta_t}) is replaced by $\Delta_t = (0.5-\overline{X_t})/t$. This allows virulence to immediately adapt to high ($\upsilon=1$) or low ($\upsilon=0.5$) values.} 
Rigorous calibration of AVA settings demonstrated that values $\alpha = 0.0125$, $\mu = 0.3$, and $\tau = 0.56$ can be applied successfully in a number of diverse domains. 
In particular, it was shown that AVA can coevolve high performing sorting networks and maze navigation agents with much greater computational efficiency than archive techniques such as LAPCA \cite{Cartlidge2011}.

\section{Substitution of the Fittest}\label{sec:sf}

First introduced in \cite{AlcarazH2021}, substitution of the fittest (SF) is a domain-independent technique designed to keep coevolving populations engaged. In essence, one can consider that SF acts to keep populations evolving at the same pace by applying a brake on the population evolving more quickly and applying an accelerator on the population evolving more slowly. 

In competitive coevolution, subjective fitness $\psi_i$ of individual $i$ is evaluated through competition such that $0\leq\psi_i\leq1$ is the proportion of victories gained by $i$. Let   $0\leq\sigma_{A}\leq1$ and $0\leq\sigma_{B}\leq1$ represent the mean subjective fitness of individuals in populations $A$ and $B$, respectively. Then we can calculate disengagement in a two-population coevolutionary system as $\delta = \lvert \sigma_{A} - \sigma_{B} \rvert$, where $0 \leq \delta \leq 1$.
 When $\delta=1$ populations are fully disengaged, such that all $n$ individuals in one population are victorious against all opponents in the other population. Each generation, $\delta$ is calculated and if it has increased from the previous generation then SF is triggered. 
 
The first step of SF is to calculate the number of individuals to be substituted as $\kappa = \lceil n\delta^{\frac{1}{\delta}} \rceil$. In the second step, we use the following rules for each population, depending on which has the lowest/highest mean relative fitness $\sigma$:

\vspace{0.1cm}
\begin{description}

    \item [Population with lowest $\sigma$:] Rank all individuals by subjective fitness $\psi_i$. Replace bottom $\kappa$ individuals with the {\em lowest} $\psi_i$ with top $\kappa$ individuals with the {\em highest} $\psi_i$. Finally, increase the subjective fitness of every individual by $\psi_{i}'= min(\psi_{i} + \delta,1)$.
    
    \item [Population with highest $\sigma$:] Rank all individuals by subjective fitness $\psi_i$. Replace top $\kappa$ individuals with the {\em highest} $\psi_i$ with bottom $\kappa$ individuals with the {\em lowest} $\psi_i$. Finally, decrease the subjective fitness of every individual by 
     $\psi_{i}'= max(\psi_{i} - \delta,0)$.
\end{description}

For more details about the dynamics induced by SF, see \cite{AlcarazH2021}.

\section{Explorations in the ``Greater Than'' Game}\label{sec:co}

The \emph{greater than} game \cite{Watson2001} was introduced as a minimal coevolutionary substrate capable of demonstrating the pathology of disengagement. Here, we use a variation of the greater than game to explore the performance and behaviour of SF against RV and AVA. We select this domain because it is analytically tractable and simple to configure, which allows us to understand coevolutionary dynamics. The solutions of the game are not of interest in themselves.

Individuals are encoded as bit-strings of length $l$. The scalar value of each individual is calculated as the sum of the number of bits equal to one. The aim of the game is to maximise this scalar value by discovering individuals with all bits equal to one (i.e., individuals with scalar value $l$). This problem is trivial in a evolutionary setting if scalar values of individuals can be directly measured. However, the greater than game assumes that the scalar values of individuals can only be accessed by comparing two values such that the subjective fitness of individual $\alpha$ is derived from a comparison against individual $\gamma$ as $score(\alpha,\gamma)=1$ if $\alpha>\gamma$; $0.5$ if $\alpha=\gamma$; and 0 otherwise. Each individual is compared against a sample of $S$ opponents drawn from the competing population, with subjective fitness calculated as the mean score in competition against these $S$ opponents. 

To control the likelihood of disengagement, we add a mutation bias parameter $\beta$, where $0\leq\beta\leq1$. When a parent is selected to generate an offspring clone, each bit in the offspring has a probability $m$ of mutating. When mutation occurs, the bit is assigned a new value at random, with probability $\beta$ of assigning a value of 1 and probability $1-\beta$ of assigning a value of 0. Consequently, when $\beta=0.5$ mutation is {\em unbiased} there is an equal chance of assigning the bit to 1 or 0; when $\beta=0$ mutation will always assign the bit to 0; and when $\beta=1$ mutation will always assign the bit to 1. This bias parameter allows the simple game to emulate the intrinsic asymmetry exhibited by more complex domains, where it is often easier for one coevolving population to outperform the other -- for instance, when coevolving list-sorting algorithms and input lists \cite{Hillis1990}, it's much easier to be an unsorted list than an algorithm that can sort a list. When populations disengage, selection pressures are removed and populations are left to drift through mutation alone. In the greater than game, we expect drifting populations to tend towards individuals with a proportion of ones equal to $\beta$. Therefore, in the case that $l=100$ and $\beta=0.25$, a drifting population will tend towards scalar values of 25. 

In our two-population competitive set up, we label the populations as {\em hosts} and {\em parasites}. Each population has an independent bias value $\beta$. We use $\beta_h$ to label the bias value of the host population and $\beta_p$ to label the bias value of the parasite population. When bias differential is high, i.e., when the value of $\beta_p$ is much larger than the value of $\beta_h$ (or vice versa), disengagement becomes more likely as the game is much easier for the parasites (alternatively, the hosts) to succeed. By varying $\beta_p$ and $\beta_h$, we are able to control the asymmetry of domain. 

\begin{figure*}[tb!]
  \centering
     \includegraphics[width=0.45\linewidth]{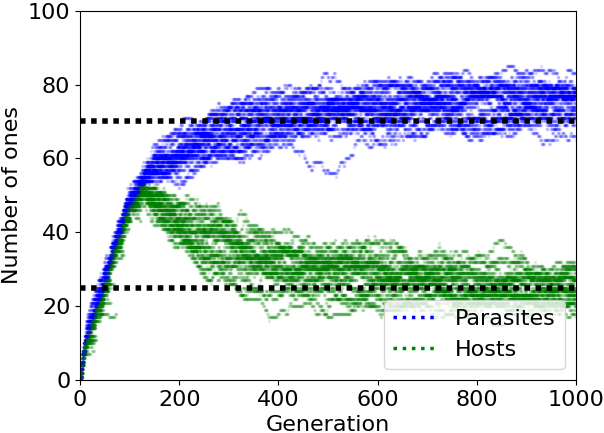}%
     \includegraphics[width=0.45\linewidth]{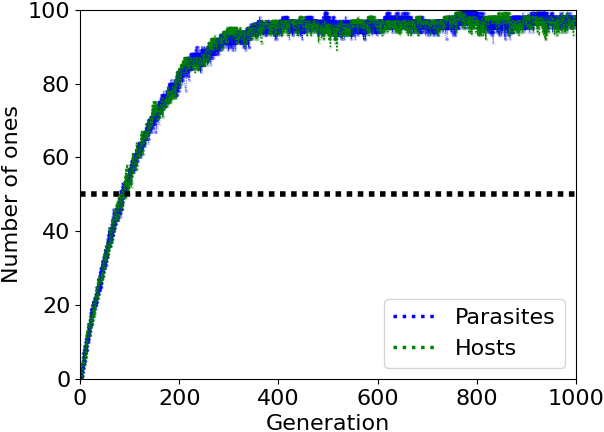}
  \caption{Example coevolutionary dynamics. On the left, when there is an inherent advantage to parasites ($\beta_p = 0.75 > \beta_h = 0.25$), populations disengage and selection pressure is removed, leaving populations to drift. On the right, when there is no advantageous bias favouring one population ($\beta_h = \beta_p = 0.5$), populations maintain engaged and selection pressure drives the system towards solutions that are close to optimum value of 100.}
  \label{fig:co_baseline}
\end{figure*}

Figure~\ref{fig:co_baseline} presents two examples of greater than coevolution to demonstrate the pathology of disengagement and the effects of bias. Scalar values of all individuals are plotted. On the left, the system has a differential bias favouring parasites, with  $\beta_p=0.75$ and $\beta_h=0.25$.
Initially, both populations remain engaged and selection drives evolutionary progress, with populations reaching scalar values of 50 by generation 150.
However, the impact of differential bias leads to a disengagement event such that all parasites are {\em greater than} their competing hosts. This results in a subjective score of zero for all hosts and a subjective score of one for all parasites. At this point selection pressure is removed as all individuals have an equal (and therefore random) chance of selection, which leaves the populations to drift under mutation alone. 
Subsequently, the high bias differential between populations ($\beta_p - \beta_h = 0.5$) ensures that disengaged populations drift through different regions of genotype space and do not re-engage through chance alone. As expected, the parasite population drifts to the parasite mutation bias (dotted line) of 75 while hosts degrade to the host mutation bias of 25.
In contrast, the figure on the right presents an example of unbiased coevolution, where populations have equal bias $\beta_p=\beta_h=0.5$.  Populations remain engaged throughout, providing a continual gradient of selection that results in the discovery of individuals with optimum scalar values equal to the maximum $l=100$. This is far higher than the scalar value of 50 (dotted line) that both populations would be expected to reach when drifting under mutation alone, i.e., when selection pressure is removed. 

\subsection{Experimental method}\label{sec:co_experimental_method}

Our experimental set up is detailed as follows. We coevolve two isolated populations, each with 25 individuals ($n=25$). The length of the binary array (an individual) is 100 ($l=100$) and each bit is initialised to 0. To generate a subjective fitness, each individual is compared against a random sample of five ($S=5$) individuals chosen from the competing population. We use tournament selection with tournament size two. Populations are asexual (i.e., there is mutation but no recombination). The probability of mutation per bit is 0.005 ($m=0.005$). Finally, each evolutionary run lasts for 1000 generations.

\subsection{Comparing RV, AVA and SF}\label{sec:co_results}

In order to measure the performance of SF, we perform a thorough comparison against RV and AVA in the greater than game. To understand how the three approaches are likely to perform in more complex domains, where populations are likely to exhibit asymmetries, we vary mutation bias across all possible levels $\beta_p,\beta_h\in[0.1, 1.0]$ s.t. $\beta_p \geq \beta_h$. 
We have assumed that hosts are the population of interest and configured mutation bias in favour of parasites (except when $\beta_p = \beta_h$); however, this choice is arbitrary as biases in favour of hosts would yield symmetrically similar results. For each bias scenario, we performed 100 experimental trials.

To analyse performance of SF, AVA and RV we utilised three metrics: (i) the reliability of the technique to maintain population engagement; (ii) the capacity to discover optimal hosts containing all ones; and (iii) the mean number of ones that hosts reach before disengagement occurs. The following sections describe our insights.

\subsubsection{Maintaining engagement}\label{sec:co_reliability}
\begin{figure*}[tb!]
  \centering
  \includegraphics[width=0.33\linewidth]{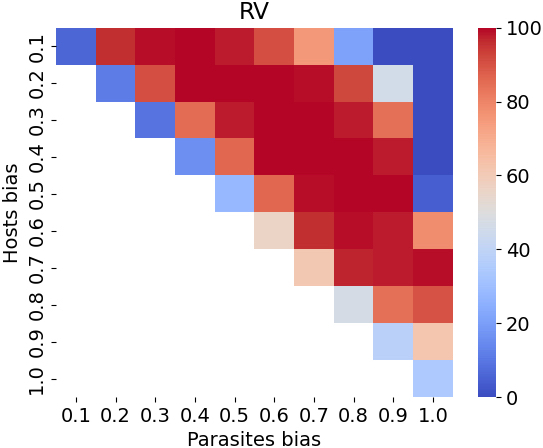}%
  \includegraphics[width=0.33\linewidth]{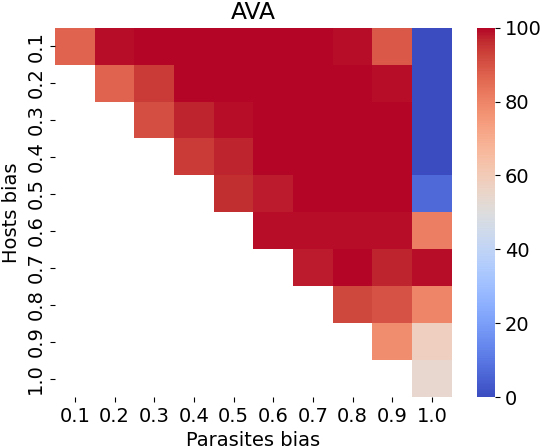}%
  \includegraphics[width=0.33\linewidth]{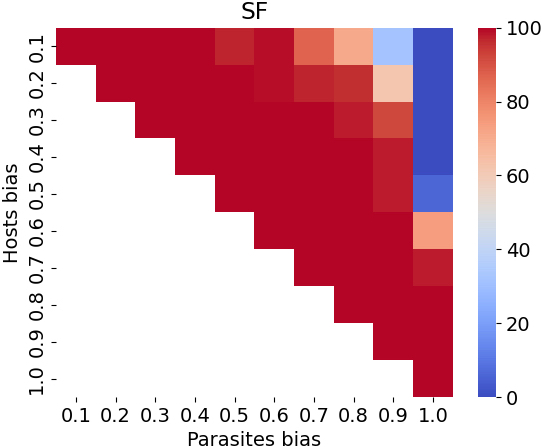}
  \centering
  \caption{Number of runs with no disengagement; RV (left), AVA (centre) and SF (right) across all bias levels (100 trials).}
  \label{fig:co_heatmaps_engagement}
\end{figure*}

The main objective of RV, AVA and SF is to maintain engagement regardless of asymmetrical bias, therefore we begin by studying the response of each technique under different bias levels.  
Figure~\ref{fig:co_heatmaps_engagement} presents a heatmap showing the number of runs (out of a total of 100 trials) where RV, AVA and SF maintained population engagement during the full coevolutionary run (regardless of whether or not an optimal host is found).  
Overall, we see that SF is able to maintain engagement more successfully than RV and AVA. For 35 of the 55 bias pairings, SF maintains engagement for the full coevolutionary run of 1000 generations across all 100 trials. For RV, however, this number is only 11, whereas for AVA the number is 22. Moreover, SF maintains engagement for the full coevolutionary run in at least 90 out of 100 trials across 45 bias pairings, while for RV this number is 29 and for AVA is 42.  
Results suggest that RV and AVA tend to struggle in scenarios (a) where the bias of both populations are either the same (symmetrical systems) or similar; and (b) where parasite bias is very high (e.g., $\beta_p=1.0$) and there is a large bias differential between parasites and hosts. In comparison, although SF also fails in scenarios where parasites have very high bias, SF is capable of maintaining engagement where RV and AVA are not.    

These results for AVA are different to those presented in the original AVA study \cite{Cartlidge2011}. We suspect two main reasons for this: (i) AVA was originally calibrated to handle bias levels across the smaller range $\beta_{h},\beta_p\in[0.5, 1.0]$; and (ii) the original experiments in \cite{Cartlidge2011} had shorter trials that ended after 750 generations. The duration of experimental trials is a key factor inasmuch AVA, in a number of bias scenarios, tends to allow disengagement {\em after} optimal hosts are found. For instance, when $\beta_h = 0.5, \beta_p=1.0$, populations tend to first reach the optimum, but then later, around generations 850-900, the populations disengage. This unexpected behaviour suggests that AVA parameters may require recalibration to maintain engagement over long time periods and over more extreme bias differentials. This demonstrates an advantage of SF, as there are no parameter settings to calibrate. 

\subsubsection{Reaching the optimum}\label{sec:co_reach_100}

\begin{figure*}[tb!]
  \centering
  \includegraphics[width=0.33\linewidth]{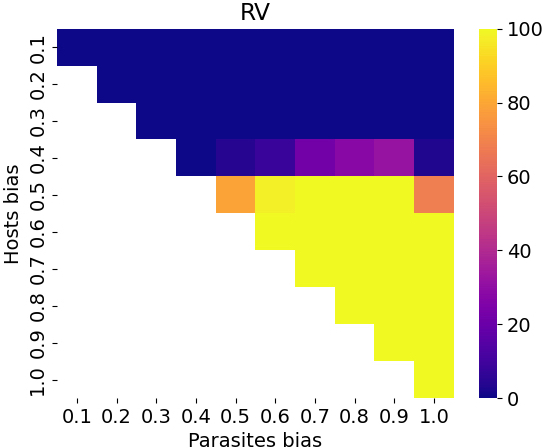}%
  \includegraphics[width=0.33\linewidth]{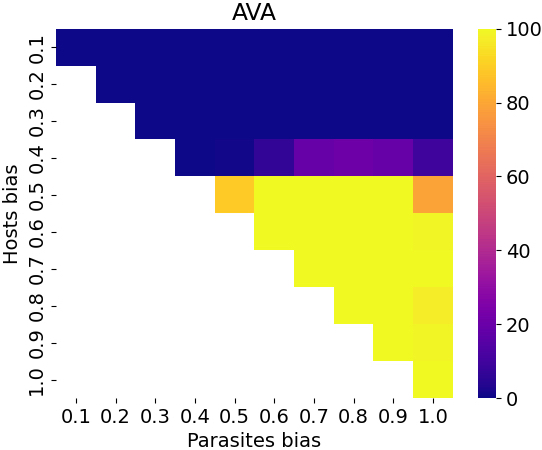}%
  \includegraphics[width=0.33\linewidth]{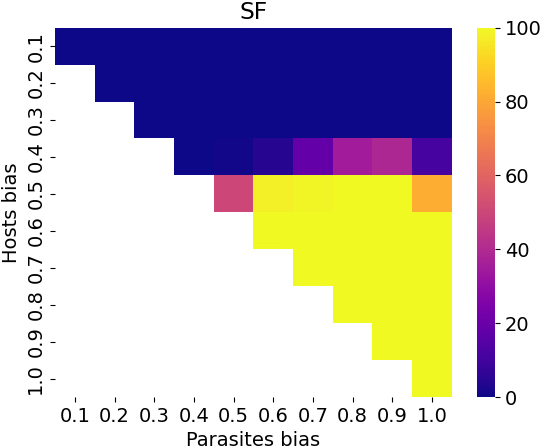}
  \centering
  \caption{Number of runs where hosts reached optimum; RV (left), AVA (centre) and SF (right) across all bias levels (100 trials).}
  \label{fig:co_heatmaps_reach_100}
\end{figure*}

Another essential aspect to analyse is the capability to reach the optimal zone. Figure~\ref{fig:co_heatmaps_reach_100} presents the number of runs where hosts (more precisely, at least one host) reached the optimal, regardless of whether or not populations disengage after this point. We see a similar pattern for RV, AVA and SF. 
RV reached the optimal zone across all 100 trials under 18 bias pairings, AVA reached it under 16 bias pairings, and SF reached it under 17 bias pairings. Furthermore, RV, AVA and SF all reached the optimal zone at least 90 times under 19 bias pairings. Perhaps unsurprisingly, for all techniques hosts were not capable of reaching the optimum when hosts have a very low mutation bias ($\beta_h < 0.5$).

\subsubsection{General performance}\label{sec:co_general_performance}

\begin{figure*}[tb!]
  \centering
  \includegraphics[width=0.33\linewidth]{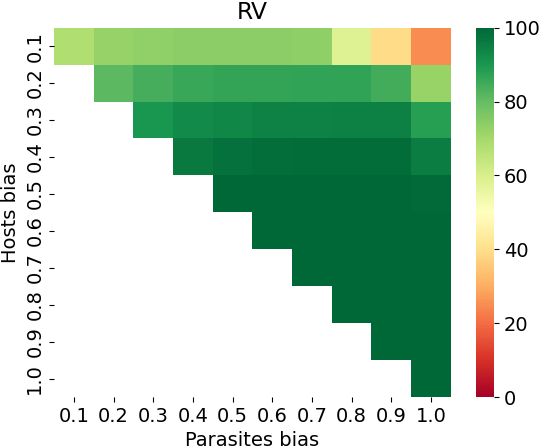}%
  \includegraphics[width=0.33\linewidth]{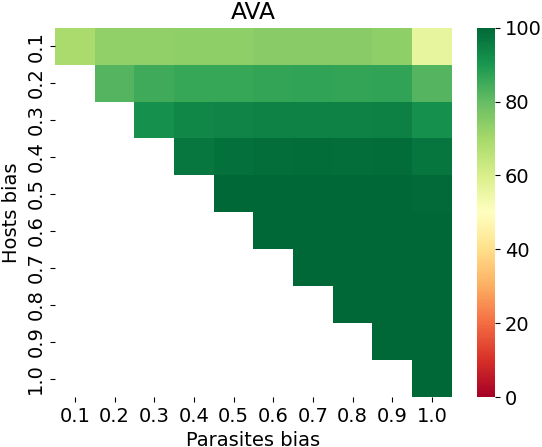}%
  \includegraphics[width=0.33\linewidth]{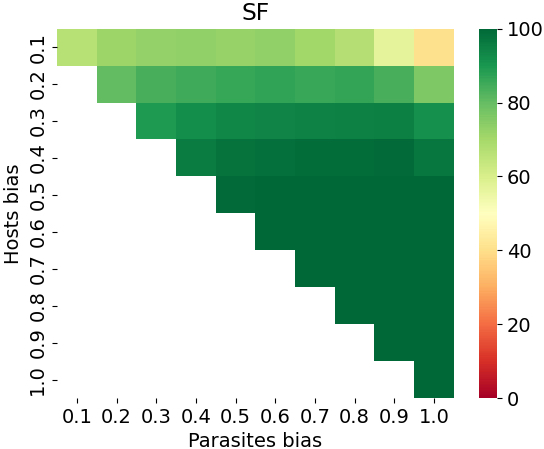}
  \centering
  \caption{Mean number of ones of best host; RV (left), AVA (centre) and SF (right) across all bias levels (100 trials).}
  \label{fig:co_heatmaps_max_ones}
\end{figure*}

Figure~\ref{fig:co_heatmaps_max_ones} shows the mean maximum number of ones of the best host across all bias configurations, regardless of whether or not populations disengage or hosts reach the optimum. 
Again, the performance of RV, AVA and SF show a similar trend.
RV reaches at least 90 ones under 34 bias scenarios, AVA reaches at least 90 ones in 36 scenarios, and SF reaches at least 90 ones under 35 scenarios. Furthermore, RV reaches 100 ones under 18 scenarios, AVA reaches 100 ones under 16 scenarios, and SF reaches 100 ones under 17 scenarios. 
These results suggest RV, AVA and SF tend to behave similarly across most bias levels. However, when there is a significant bias differential  (e.g., $\beta_h=0.1,\beta_p=0.9$), AVA enables populations to reach a greater performance (closer to the optimum) than both RV and SF. 

\subsubsection{Summary}\label{sec:co_summary}

Overall, SF has demonstrated reliable capacity to overcome disengagement in the simple {\em greater than} game across a wide range of asymmetrical bias differentials. However, the performance of SF is not significantly better than AVA or RV. Yet, unlike RV and AVA, SF offers the advantage of having no tunable parameters and therefore requires no domain calibration. This suggests that SF may be more reliable in complex domains where asymmetry is {\em apriori} unknown and likely to vary over time. In the following section, we test this hypothesis by exploring the performance of the three techniques in a more complex domain with potential real-world application.

\section{Coevolving Well-being Recommendations}\label{sec:cers}

In this section we explore the performance of SF, RV and AVA in the domain of recommender systems for health and well-being. Furthermore, we introduce a coevolutionary adaptation of an evolutionary recommender system, first introduced in \cite{AlcarazH2022}. 

\begin{figure}[tbp]
\centerline{\includegraphics[width=0.65\linewidth]{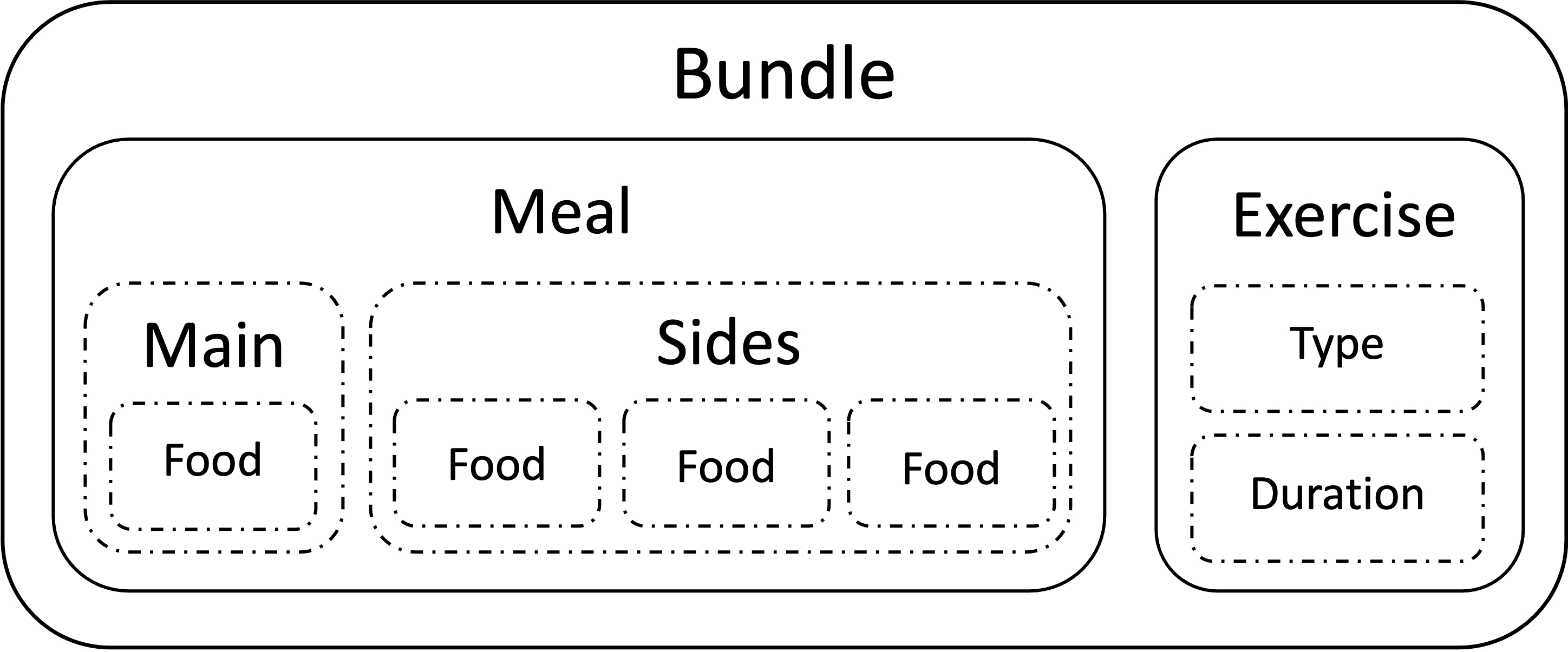}}
\caption{An individual contains three bundles, each consisting of one meal and one exercise.}
\label{fig:cers_bundle}
\end{figure}

\subsection{EvoRecSys: Evolutionary Recommender System}\label{sec:cers_original_model}

EvoRecSys (Evolutionary Recommender System) is a multi-objective health recommender system that uses a genetic algorithm to discover optimal recommendations. The constraints on recommendations are defined by user’s preferences, their well-being goal, and their physical condition. The output of EvoRecSys contains a list of bundles enclosing a meal plan (a set of ingredients) and an exercise activity (see Figure~\ref{fig:cers_bundle} for a schematic of a bundle). 
Each food item contains: (i) name; (ii) type of food; (iii) suitability for vegetarians/vegans; (iv) nutritional information (in grams) - serving size, protein, carbohydrates, sugar, fibre, fat, saturated fat, sodium; and (v) kilocalories associated with the serving size. In addition, the exercise activity contains: (i) name; (ii) intensity; (iii) Metabolic Equivalent of Task (MET); and (iv) duration.

Under the evolutionary approach, individuals are represented by a list of three bundles. The fitness $\phi_{i}$ of individual $i$ is calculated as the error across four restrictions:

\begin{equation}\label{eq:fitness}
    \phi_{i} = \frac {hf_i + ea_i + cd_i + \Psi_i} {4}
\end{equation}
\vspace{0.1cm}

\noindent
where $hf_i$ (healthy food) evaluates the amount of nutrients contained in a food item, following the dietary recommendations published by Public Health England \cite{England2016}; $ea_i$ (exercise activity) evaluates the degree of matching between a recommended exercise time and the usual time that the user spends exercising; $cd_i$ (consistency and diversity) evaluates the diversity and proportionality of meal items and exercise; and $\Psi_i$ (user preferences) evaluates how likeable the recommended items are in comparison to the user’s preferences. The error is the average of these restrictions, normalised to the range $[0.0,1.0]$. 

Tournament selection (size two) is used to select individuals for reproduction, where the fittest individual is the one with the lowest error $\phi_{i}$. A child offspring is first created as an exact clone of parent $F$. Then, the genetic operators act on the child as follows:

\begin{description}
\item[Crossover:] Occurs with probability $P_c$. A second parent $M$ is selected at {\em random} from the population. Then, some elements from parent $M$ are injected into the child following the method of \cite[Algorithms 8-9]{AlcarazH2022}. Each bundle in the child has items injected from parent $M$ with  probability $P_b=0.9$. A single item of the bundle is injected into the child such that main food is injected with probability 0.2, side food is injected with probability 0.6, and exercise activity is injected with probability 0.2. 

\item[Mutation:] Occurs with probability $P_m$. Collaborative filtering is used to find the nearest neighbour $N$ with most similar user preferences to parent $F$. One element in the child is then selected at random with uniform probability and is replaced by the equivalent element taken from the most similar neighbour $N$. This non-standard mutation operator is described in detail in \cite[Algorithms 10-11]{AlcarazH2022}.

\end{description}
\vspace{0.1cm}

These relatively complex genetic operators have been shown to improve performance of EvoRecSys \cite{AlcarazH2022-cec}.
Table~\ref{tab:fitness_values} presents four illustrative examples of individual bundles that have evolved during one run of EvoRecSys. We see that a relatively small change in fitness error of a bundle can make a large qualitative difference in the output. The bottom row in the table has coherent output that meets user preferences with an acceptable level. Therefore, we consider \textbf{0.33} as a suitable fitness threshold (i.e., the highest {\em acceptable} error). 

\begin{table}[tb!]
\caption{Fitness (error) values of example bundles for a non-vegetarian user with 2103 daily calories intake
(701/meal), 90 minutes per exercise session, and goal of “losing weight”.}
\label{tab:fitness_values}
\footnotesize{
\begin{center}
\renewcommand{\arraystretch}{1.3}

\ifnum\PREPRINT=0
    \begin{tabular}{ c p{2.7cm} p{1.2cm} p{4.4cm} }
\else
    \begin{tabular}{ c p{4cm} p{2.cm} p{7cm} } 
\fi
 \toprule
 \textbf{Fitness} & \textbf{Food (g)} & \textbf{Exercise} & \textbf{Observations} \\ [0.5ex]
 \midrule
 0.6739 & mussels (140.7)\newline eggplant (80.8)\newline lettuce (806.3)\newline eggplant (1978.6) & Football\newline(117 min) & Abnormal serving size ($>2$kg of eggplant) and repeated food items. Exercise time $30\%$ too long.\\ 
 
 0.4395 & scrambled egg (159.4) \newline leek (760.2) \newline almond (3.3) \newline crackers (1.9) &  Tai-Chi\newline(86 min)    & Uneven serving sizes (smallest $<2$g).  Exercise time is close to target of 90 min. \\ 
 
 0.3864 & pollock (230.8)\newline lettuce (209.7)\newline leek (333.2)\newline potato (115.1) & Yoga\newline(78 min)    & Serving sizes balanced. Exercise time is 13\% lower than target time. \\
 
 0.3297 & rice noodles (241.8)\newline bean sprout (258)\newline almond (11.4)\newline leek (158.5) & Pilates\newline(89 min) & Serving sizes are consistent. Exercise time is within 2\% of target. \\ 
 \bottomrule
\end{tabular}
\end{center}
}
\end{table}

\subsection{Coevolution: CoEvoRecSys}\label{sec:cers_coevolution}
One potential use of health recommender systems consists of building weekly or monthly meal-activity plans. Since EvoRecSys has been designed to provide recommendations for a single day, there is a tendency towards a lack of diversity in recommendations over time. To address this issue, we introduce a coevolutionary adaptation, which we name CoEvoRecSys. The aim of CoEvoRecSys is to generate more diverse meal-activity that nevertheless still ensure plans meet user preferences and well-being goals. 

In order to generate a coevolutionary implementation of EvoRecSys, we use two competing populations, {\em hosts} and {\em parasites}. Parasites are initialised using user's physical data (to calculate the suggested number of intake calories according to well-being goal) and food and exercise preferences (e.g., if the user is vegetarian and cannot swim, a recommendation of a beefsteak meal followed by a session of swimming is unlikely to appear). This is the canonical initialisation method utilised in \cite{AlcarazH2022}. In contrast, hosts are initialised at random: neither user's physical data nor user's preferences are considered. We include this host population to introduce diversity. This initialisation routine ensures that parasites have an intrinsic fitness advantage over hosts as parasites are much more likely to meet user preferences and therefore have lower error $\phi_{i}$ than randomly initialised parasites. 

A subjective fitness score for individuals is calculated by comparing error values of individual $i$ against a sample of $S$ opponents from the other population. Comparative score of host $i$ is defined as:

\begin{equation}\label{eq:cers_sub_apt}
    score(\phi_{h_i},\phi_{p_j}) = 
    \begin{cases}
      1 & \text{if $\phi_{h_i} <  \phi_{p_j}$} \\
      0.5 & \text{if $\phi_{h_i} = \phi_{p_j}$} \\
      0 & \text{otherwise}
    \end{cases} 
\end{equation}
\vspace{0.1cm}

\noindent
where $\phi_{h_i}$ is the error of host $i$ and $\phi_{p_j}$ is the error of parasite $j$ (see Equation~\ref{eq:fitness}). Subjective fitness of host $i$ is then calculated as the mean score against $S$ parasite competitors drawn at random from the other population.

\subsection{Experimental Configuration}
Our experimental set-up consists of two coevolving populations. We use a sample size $S=5$ to obtain the competitive fitness of individuals. Tournament selection is used with tournament size two, and mutation and crossover operators are implemented for both populations with probabilities $P_c=0.8$ and $P_m=0.1$, respectively. Each run lasts 500 generations. For user preferences and physical attributes, we re-use the anonymised data collected from participants during the original EvoRecSys experiments \cite{AlcarazH2022}.

\begin{figure*}[tb!]
  \centering
     \includegraphics[width=0.45\linewidth]{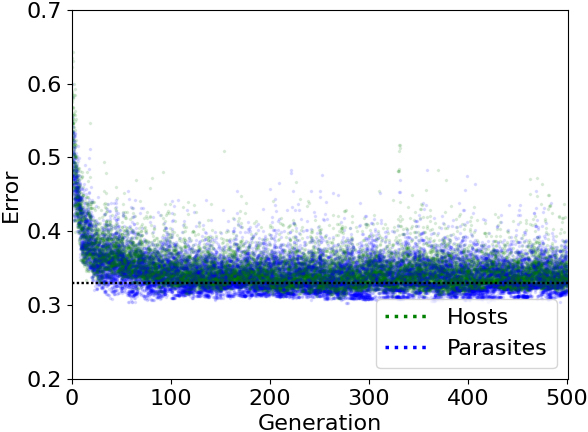}
     \includegraphics[width=0.45\linewidth]{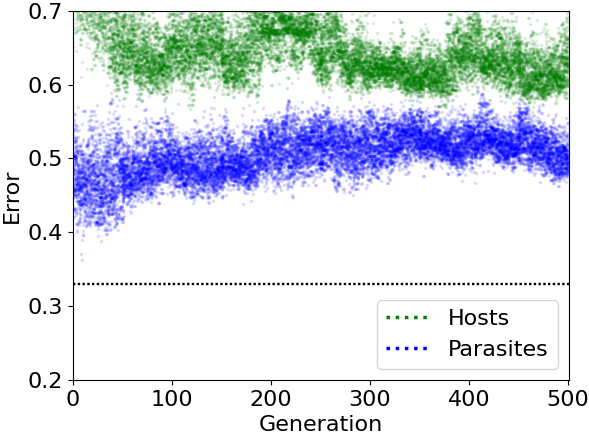}%
  \caption{CoEvoRecSys examples: When populations are engaged (left), selection pressure encourages the discovery of {\em acceptable} solutions that reach the maximum error threshold of 0.33. When populations disengage (right), parasites easily outperform hosts and the system tends to stabilise in a mediocre state with error well above the acceptable threshold.}
  \label{fig:cers_baseline}
\end{figure*}

\subsection{Disengagement in CoEvoRecSys}\label{cers_disengagement}

Figure~\ref{fig:cers_baseline} presents results from two example runs of CoEvoRecSys. On the left, we see an example where host and parasite populations remain engaged throughout the run, with both populations obtaining acceptable solutions (i.e., populations reach the recommendation threshold). On the right, we see an example where hosts and parasites quickly disengage, leaving the populations to drift at random through genetic space. From the start, parasites easily outperform hosts due to their inherent advantage of incorporating user preferences during initialisation. The system falls into a mediocre stable state (e.g., see \cite{Ficici98a}) such that progress halts and acceptable recommendations are never discovered. Therefore, to ensure CoEvoRecSys is able to discover acceptable solutions, it is imperative to avoid disengagement.

\subsection{CoEvoRecSys: Comparison of RV, AVA and SF}\label{sec:cers_results}
In this section, we compare the ability of RV, AVA and SF to counteract disengagement in CoEvoRecSys. As a baseline, also compare the performance of CoEvoRecSys with no disengagement-mitigation technique applied.
We trial each approach in simulations where we vary the size of each population $n\in\{30,60,130,260,510\}$. Under all conditions, we perform 30 experimental trials. We configure RV and AVA using settings taken from their original publications. Hence, we set RV to have virulence $v=0.75$ \cite{Cartlidge2004b}; for AVA, we set learning rate $\alpha = 0.0125$, momentum $\mu = 0.3$, and target value $\tau = 0.56$ \cite{Cartlidge2011}. 

To analyse performance of RV, AVA and SF, four metrics are used: (i) capacity to maintain engagement (Section~\ref{sec:cers_reliability}); (ii) capacity to discover acceptable recommendations (Section~\ref{sec:cers_general_performance}); (iii) capacity to discover diverse recommendations (Section~\ref{sec:cers_div_performance}); and (iv) capacity to discover both acceptable and diverse recommendations  (Section~\ref{sec:cers_multi_objective}).  

\begin{figure*}[tb!]
  \centering
  \includegraphics[width=0.40\linewidth]{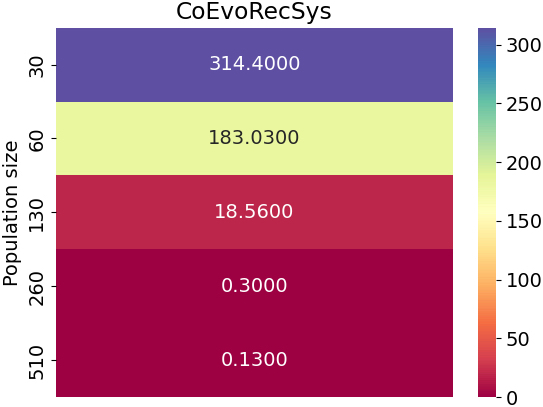}
  \includegraphics[width=0.40\linewidth]{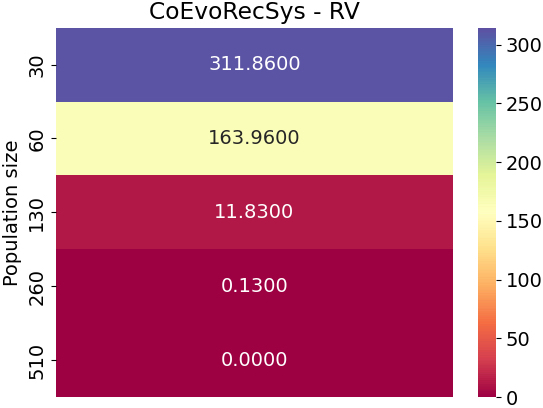}
  \includegraphics[width=0.40\linewidth]{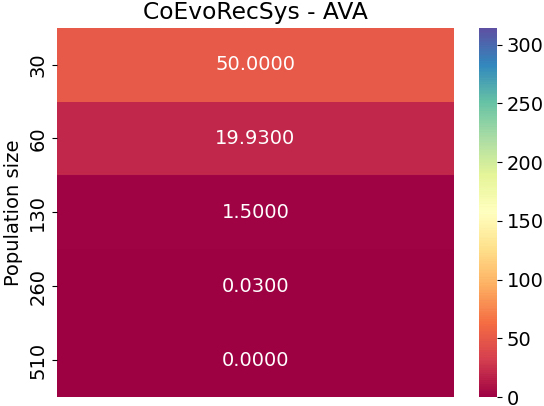}
  \includegraphics[width=0.40\linewidth]{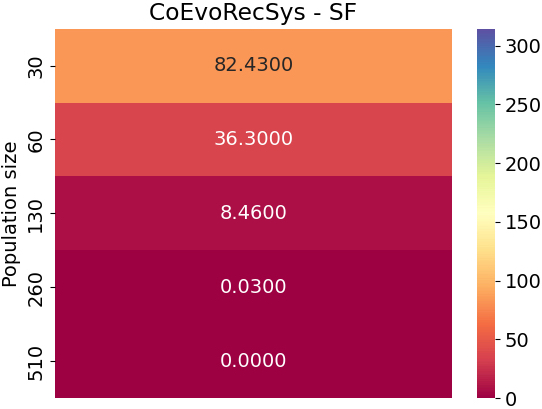}
  \centering
  \caption{Mean number of disengaged generations during an experimental trial; CoEvoRecSys (top-left); CoEvoRecSys with RV (top-right), CoEvoRecSys with AVA (bottom-left) and CoEvoRecSys with SF (bottom-right) across all parameter configurations (30 trials).}
  \label{fig:cers_reliability}
\end{figure*}

\subsubsection{Ability to maintain engagement}\label{sec:cers_reliability}
We begin by exploring the capacity of RV, AVA and SF to maintain engagement in CoEvoRecSys. Figure~\ref{fig:cers_reliability} presents heatmaps illustrating the mean number of disengaged generations that occur during the full evolutionary process across all combinations of $n$ (number of individuals in each population). 

In general, as population size increases (top to bottom of each heatmap), we see that the number of disengaged generations falls. This is to be expected, as larger populations mean that there is a greater chance for populations to engage through random mutation of individuals. When population size is large (i.e., when $n\in\{260,510\}$), there is very little disengagement under any condition, including the baseline. However, for smaller population sizes (i.e., when $n\in\{30,60\}$), disengagement is common in the baseline CoEvoRecSys. The results for RV are similar to the baseline, indicating that RV may require recalibration. In contrast, AVA and SF are able to counter the effects of disengagement, resulting in many fewer disengaged generations when population sizes are small. While AVA and SF both perform well overall, results suggest that AVA is able to maintain engagement more reliably than SF.

\begin{figure*}[tb!]
  \centering
  \includegraphics[width=0.40\linewidth]{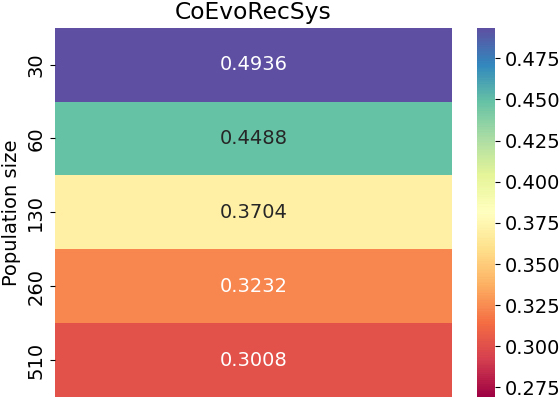}
  \includegraphics[width=0.40\linewidth]{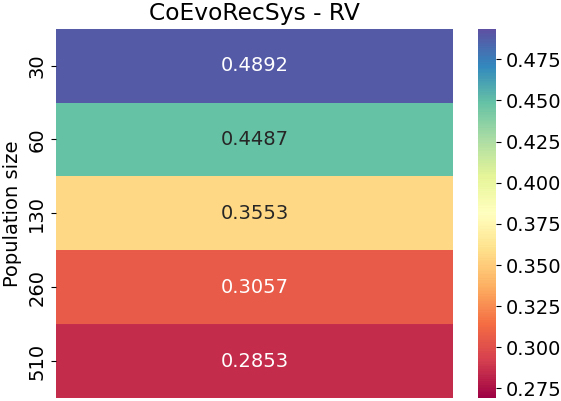}
  \includegraphics[width=0.40\linewidth]{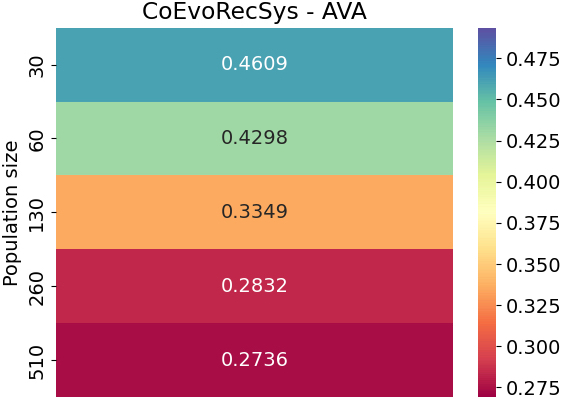}
  \includegraphics[width=0.40\linewidth]{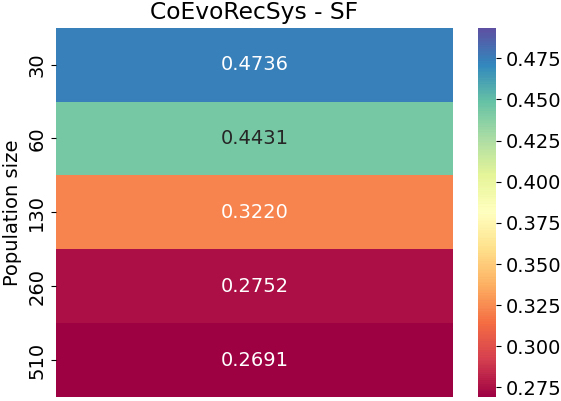}
  \centering
  \caption{Mean error of best host: CoEvoRecSys (top-left); CoEvoRecSys under RV (top-right); CoEvoRecSys under AVA (bottom-left); and CoEvoRecSys underSF (bottom-right).}
  \label{fig:cers_general_performance}
\end{figure*}

\subsubsection{Ability to find acceptable recommendations}\label{sec:cers_general_performance}
One of the most important metrics of performance is the ability to discover recommendations that have error below the acceptable threshold of 0.33.  Figure~\ref{fig:cers_general_performance} presents heatmaps showing the mean error reached by the best host during each evolutionary run. 
As expected, under all four conditions, when the number of individuals is small the performance is poor; and performance improves as population sizes are increased.

We see that the baseline system just about manages to reach the acceptable threshold when population size is $n=260$. At this population size, RV, AVA, and SF all discover solutions of better quality than the baseline. Indeed, SF is able to find acceptable solutions when population size is only $n=130$. To determine whether differences in performance are statistically significant, we first confirm that results data are normally distributed (Shapiro-Wilk test; $p>0.05$, so cannot reject the null). The result of one-way ANOVA shows that there are significant differences among the approaches ($p<0.01$). Finally, post-hoc tests (Tukey-HSD test; $p<0.01$) determine that there are significant differences in performance between all techniques. This generates the following ranking for recommendation performance: SF $>$ AVA $>$ RV $>$ baseline.

\begin{figure*}[tb!]
  \centering
  \includegraphics[width=0.40\linewidth]{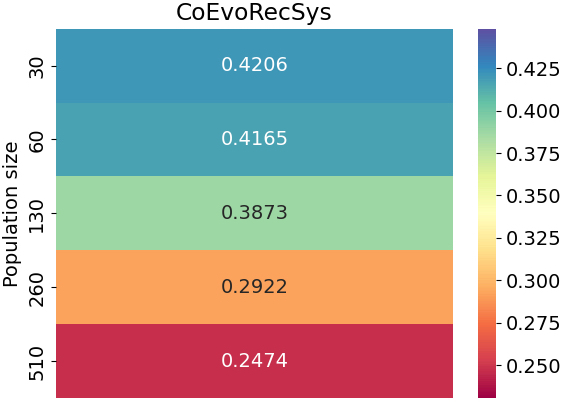}
  \includegraphics[width=0.40\linewidth]{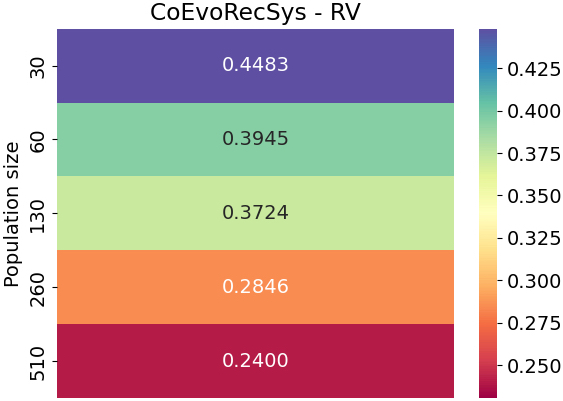}
  \includegraphics[width=0.40\linewidth]{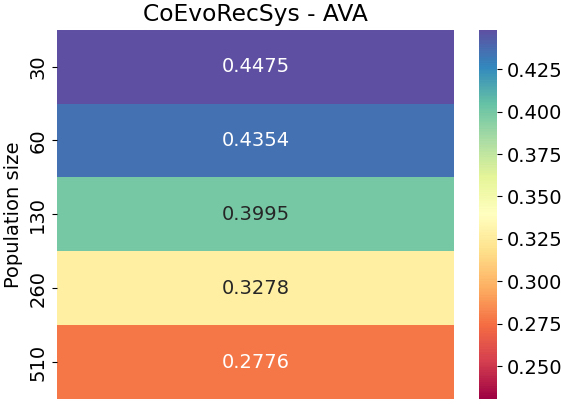}
  \includegraphics[width=0.40\linewidth]{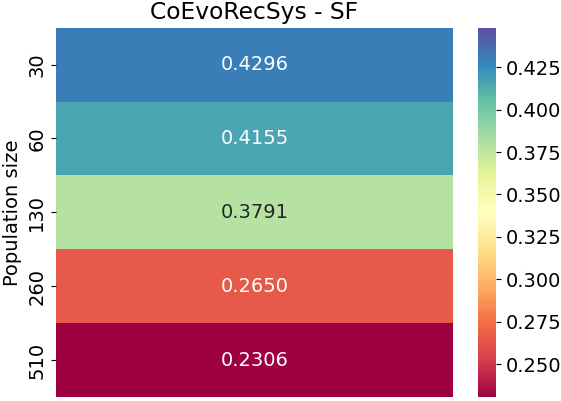}
  \centering
  \caption{Mean diversity error of best host: CoEvoRecSys (top-left); CoEvoRecSys under RV (top-right), CoEvoRecSys under AVA (bottom-left); and CoEvoRecSys SF (bottom-right).}
  \label{fig:cers_diversity}
\end{figure*}

\subsubsection{Ability to find diverse recommendations}\label{sec:cers_div_performance}
In a practical scenario, CoEvoRecSys would likely be used to provide meal plans over a longer time period than a one day. Therefore, diversity in recommendations is important so that users are not expected to eat the same meals and perform the same exercise activities every day. Here, we generate recommendations for a 28 day period (i.e., one month) and evaluate their diversity. The evaluation function is an adaptation of $cd_i$ (see Section~\ref{sec:cers_original_model}), such that diversity of item categories are considered in addition to items themselves; for example \{{\em carrots, asparagus, cauliflower}\} has a lower diversity than \{{\em beans, quinoa, broccoli}\} because all items in the former set are vegetables.

In order to provide an acceptable diversity threshold, we used EvoRecSys to conduct 28 experimental trials (equivalent to one month of recommendations). EvoRecSys was configured using the same parameter settings presented in \cite{AlcarazH2022}, apart from elitism was disabled and the length of each run was set to 500 generations. The mean diversity error was 0.2753, with standard deviation 0.022. We therefore consider \textbf{0.2753} as the maximum acceptable threshold of diversity error.

Results of running CoEvoRecSys are presented in Figure~\ref{fig:cers_diversity}. These heatmaps show the mean diversity error reached by the best host, where a {\em lower} value of diversity error indicates a {\em more} diverse set of recommendations. Once again, we observe the general trend that larger populations produce higher performance (i.e., greater diversity). 
While the heatmaps suggest that SF produces the most diversity (reaching the threshold when $n=260$) and AVA produces the least diversity (marginally failing to reach the threshold even when $n=510$), these differences are not statistically significant. We determine this by first confirming that the data is not normally distributed (Shapiro-Wilk test; $p<0.05$, so reject the null); then, using the non-parametric Kruskal-Wallis test, we are unable to reject the null hypothesis that there are no significant differences between the four approaches (Kruskal-Wallis test; $p>0.05$).

\begin{figure*}[tb!]
  \centering
  \includegraphics[width=0.40\linewidth]{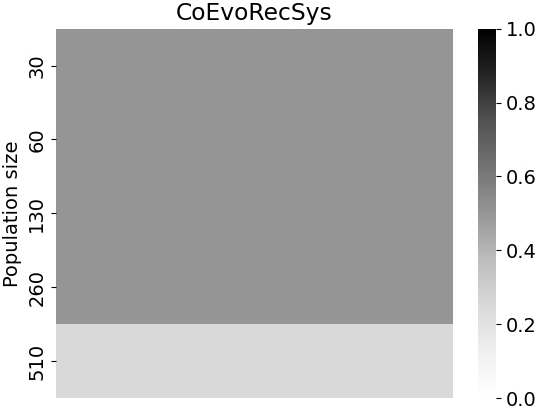}
  \includegraphics[width=0.40\linewidth]{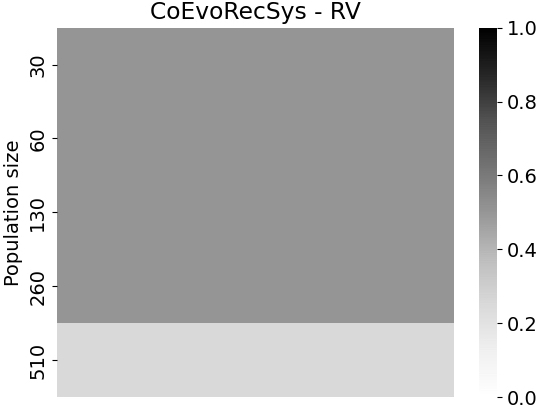}
  \includegraphics[width=0.40\linewidth]{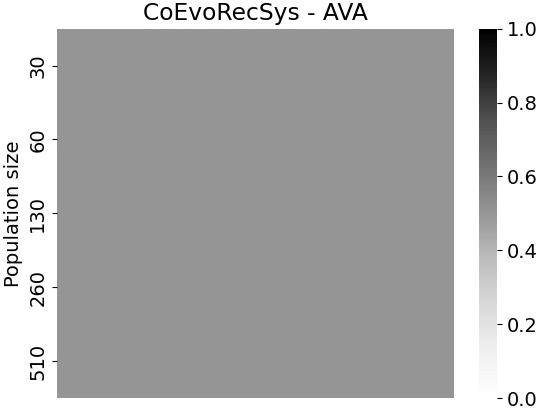}
  \includegraphics[width=0.40\linewidth]{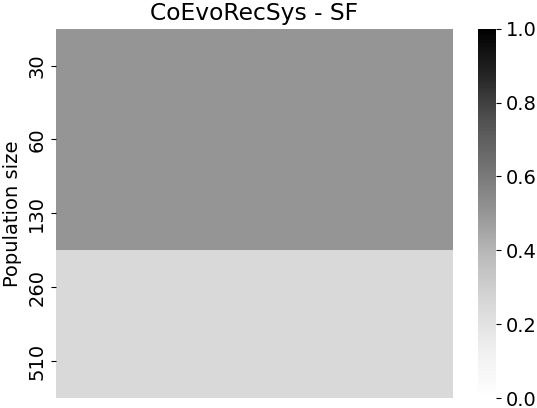}
  \centering
  \caption{Parameter configurations (light shade) that reach both fitness threshold and diversity threshold; CoEvoRecSys (top-left); CoEvoRecSys with RV (top-right); CoEvoRecSys with AVA (bottom-left); and CoEvoRecSys with SF (bottom-right).}
  \label{fig:cers_multi}
\end{figure*}

\subsubsection{Ability to find acceptable {\em and} diverse recommendations}\label{sec:cers_multi_objective}
Here, we consider configurations where solutions are {\em both} acceptable and diverse. Figure~\ref{fig:cers_multi} presents heatmaps showing configurations where both thresholds are reached, indicated by the light-grey regions. 
Results shows that AVA is not capable of reaching both thresholds in any configuration: although AVA finds high quality recommendations, it marginally misses the required threshold of diversity.  In comparison, RV and the baseline are able to reach both thresholds when population sizes are large ($n=510$). However, SF is capable of achieving both thresholds even with smaller population size $n=260$. Therefore, we conclude that SF is the most efficient and robust approach for this CoEvoRecSys problem domain. 

\begin{figure*}[tb!]
  \centering
  \includegraphics[width=0.50\linewidth]{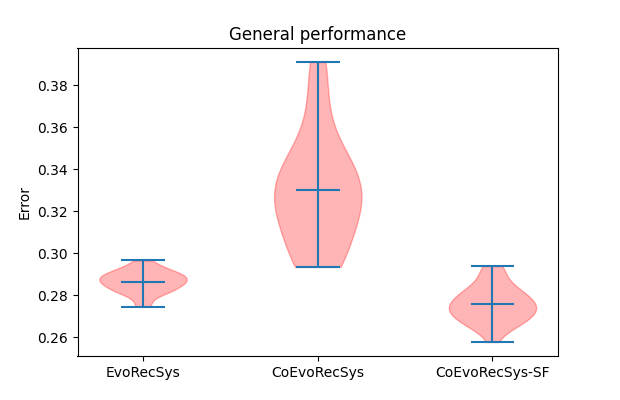}%
  \includegraphics[width=0.50\linewidth]{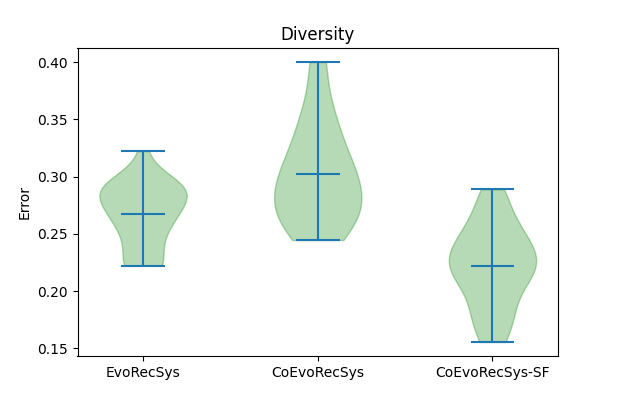}
  \centering
  \caption{Comparison of EvoRecSys and CoEvoRecSys across one month of recommendations. CoEvoRecSys with SF has best performance (left) and best diversity (right).}
  \label{fig:cers_cers_vs_ers}
\end{figure*}

\subsection{Evolution vs Coevolution}\label{sec:cers_ers_vs_cers}
Here, we perform a comparison of EvoRecSys against CoEvoRecSys over 28 experimental trials (equivalent to one month of meal-exercise recommendations). Based on the results obtained in Section~\ref{sec:cers_results}, we select SF as the best technique to use in CoEvoRecSys.
In order to perform a comparison in a real-based environment, we use the set-up implemented for the web-based application introduced in \cite{AlcarazH2022}. Table~\ref{tab:ers_web_parameters} presents the values utilised for each parameter.
We utilise a sample size of five opponents ($S=5$) to calculate competitive score. We use tournament selection with tournament size two ($T=2$) and elitism is not implemented. 

\begin{table}[tb!]
\caption{Parameter settings utilised in the web-based application introduced in \cite{AlcarazH2022}.}
\label{tab:ers_web_parameters}
\small{
\begin{center}
\begin{tabular}{ lc }
 \toprule
 \textbf{Parameter}  & \textbf{Value} \\ [0.5ex]
 \midrule
 Number of individuals & 250 \\ 
 Number of generations & 150 \\ 
 Crossover probability & 0.6 \\ 
 Mutation probability &  0.1 \\ 
 \bottomrule
\end{tabular}
\end{center}
}
\vspace{-4mm}
\end{table}

Figure~\ref{fig:cers_cers_vs_ers} presents violin plots showing comparative performance of EvoRecSys, CoEvoRecSys and CoEvoRecSys with SF. Each violin presents minimum, maximum, and median value, as well as a kernal density estimation of the frequency distribution of values across all 28 runs. 
On the left plot of general performance, we see that CoEvoRecSys with SF tends to reach recommendations that are significantly more suitable and coherent than both EvoRecSys and the CoEvoRecSys baseline (Shapiro-Wilk test: $p>0.05$; one-way ANOVA test and Tukey-HSD test: $p<0.01$). On the right plot of diversity, we see CoEvoRecSys with SF also tends to reach significantly more diverse food items and exercise activities than both EvoRecSys and the CoEvoRecSys baseline (Shapiro-Wilk test: $p<0.05$; Kruskal-Wallis test and Dunn's test: $p<0.01$).
 
In general, results indicate that the coevolutionary approach with SF significantly outperforms the evolutionary approach of EvoRecSys in terms of suitable and diverse recommendations. When coevolution is used without SF, strong asymmetrical bias between host and parasite populations produces disengagement, which leads to detrimental performance. By mitigating against disengagemnet, SF enables the coevolutionary system to maintain engagement and discover solutions that are better than those discovered by single-population evolution alone.

\section{Conclusions}\label{sec:conclusions}
We have introduced and explored the capacity of {\em substitution of the fittest} (SF) to counteract disengagement, a pathology that hinders progress in two-population competitive coevolutionary GAs. First, using the deliberately simple {\em greater than} domain, SF was shown to overcome disengagement across a wide range of asymmetrical bias differentials (see Section~\ref{sec:co}). When compared against RV and AVA, alternative techniques from the literature, SF does not have significantly better performance overall. However, unlike RV and AVA, SF offers the advantage of having no tunable parameters and therefore requires no domain calibration. This suggests that SF may be more reliable in complex domains where asymmetry is {\em apriori} unknown and likely to vary over time. 

Subsequently, we attempted the more challenging real-world problem of coevolving recommendations for health and well-being (see Section~\ref{sec:cers}). We called this coevolutionary approach {\em CoEvoRecSys} as it extends a previously published evolutionary approach to recommender systems called {\em EvoRecSys}.  
Results showed that the baseline CoEvoRecSys suffers from disengagement, resulting in poorer solutions than those discovered by EvoRecSys. However, when SF is introduced into CoEvoRecSys, disengagement is successfully countered, leading to solutions that are significantly better than those discovered by EvoRecSys. 
We also show that SF outperforms both RV and AVA in CoEvoRecSys. However, for these experiments, we configure RV and AVA using parameter values taken from the literature and it is therefore possible that RV and AVA underperform because their parameter values require recalibration. In contrast, SF has no parameters to calibrate, making it a more robust domain-independent technique that can be used ``out of the box''.

In future, we plan to extend CoEvoRecSys as a web-service or mobile application that enables users to interact with the system to provide real-time feedback on the subjective quality of recommendations. We will also further explore the performace of SF in coevolutionary domains with natural asymmetry between populations, such as maze-robot navigation or list-sorting networks (e.g., see \cite{Cartlidge2011}).

\section*{Declarations}


\noindent
{\bf Funding:} Hugo Alcaraz-Herrera's PhD is supported by The Mexican Council of Science and Technology (Consejo Nacional de Ciencia y Tecnología - CONACyT). 

\vspace{0.2cm}

\noindent
{\bf Conflict of interest:} The authors have no conflicts of interest. 

\vspace{0.2cm}

\noindent
{\bf Ethics approval:} This article does not contain any studies with human participants or animals performed by any of the authors.

\vspace{0.2cm}

\noindent
{\bf Informed consent:}  Participants' anonymous preference data used for building health and well-being recommendations in Section~\ref{sec:cers} was previously collected for the work on EvoRecSys, published in \cite{AlcarazH2022}. Informed consent was obtained from all individual participants included in the EvoRecSys study.

\ifnum\PREPRINT=1
    \bibliographystyle{splncs03}
\fi

\bibliography{sfe_bibliography}

\begin{thebibliography}{10}
\providecommand{\url}[1]{\texttt{#1}}
\providecommand{\urlprefix}{URL }

\bibitem{Akinola2020}
Akinola, A., Wineberg, M.: Using implicit multi-objectives properties to
  mitigate against forgetfulness in coevolutionary algorithms. In: Genetic and
  Evolutionary Computation Conference. pp. 769--777. GECCO (2020)

\bibitem{AlcarazH2021}
Alcaraz-Herrera, H., Cartlidge, J.: Substitution of the fittest: A novel
  approach for mitigating disengagement in coevolutionary genetic algorithms.
  In: Proceedings of the 13th International Joint Conference on Computational
  Intelligence. pp. 59--67. ECTA (2021)

\bibitem{AlcarazH2022-cec}
Alcaraz-Herrera, H., Cartlidge, J.: Exploration of ontological representations
  for evolutionary computation. In: IEEE Congress on Evolutionary Computation.
  pp. 1--8. CEC (2022)

\bibitem{AlcarazH2022}
Alcaraz-Herrera, H., Cartlidge, J., Toumpakari, Z., Western, M., Palomares, I.:
  {EvoRecSys}: Evolutionary framework for health and well-being recommender
  systems. User Modeling and User-Adapted Interaction  (January 2022)

\bibitem{Bari2018}
Bari, A.G., Gaspar, A., Wiegand, R.P., Bucci, A.: Selection methods to relax
  strict acceptance condition in test-based coevolution. In: Congress on
  Evolutionary Computation. pp. 1--8. CEC (2018)

\bibitem{Bullock2002}
Bullock, S., Cartlidge, J., Thompson, M.: Prospects for computational steering
  of evolutionary computation. In: Workshop Proceedings 8th Int. Conf. on
  Artificial Life. pp. 131--137. ALIFE (Dec 2002),
  \url{https://eprints.soton.ac.uk/261459/1/Prospects.pdf}

\bibitem{Cartlidge2002}
Cartlidge, J., Bullock, S.: Learning lessons from the common cold: How reducing
  parasite virulence improves coevolutionary optimization. In: Congress on
  Evolutionary Computation. pp. 1420--1425 vol.2. CEC (2002)

\bibitem{Cartlidge2011}
Cartlidge, J., Ait-Boudaoud, D.: Autonomous virulence adaptation improves
  coevolutionary optimization. IEEE Transactions on Evolutionary Computation
  15(2),  215--229 (April 2011)

\bibitem{Cartlidge2003}
Cartlidge, J., Bullock, S.: Caring versus sharing: How to maintain engagement
  and diversity in coevolving populations. In: Banzhaf, W., Ziegler, J.,
  Christaller, T., Dittrich, P., Kim, J.T. (eds.) Advances in Artificial Life.
  pp. 299--308. Springer Berlin Heidelberg, Berlin, Heidelberg (2003)

\bibitem{Cartlidge2004b}
Cartlidge, J., Bullock, S.: Combating coevolutionary disengagement by reducing
  parasite virulence. Evolutionary Computation  12(2),  193--222 (June 2004)

\bibitem{Cartlidge2004a}
Cartlidge, J., Bullock, S.: Unpicking tartan {CIAO} plots: Understanding
  irregular coevolutionary cycling. Adaptive Behavior  12(2),  69--92 (2004)

\bibitem{DeJong2007}
{de Jong}, E.D.: A monotonic archive for pareto-coevolution. Evolutionary
  Computation  15(1),  61--93 (Mar 2007)

\bibitem{Ficici98a}
Ficici, S.G., Pollack, J.B.: Challenges in coevolutionary learning: Arms-race
  dynamics, open-endedness, and mediocre stable states. In: International
  Conference on Artificial Life. pp. 238--247. ALIFE (1998),
  \url{https://dl.acm.org/doi/10.5555/286139.286166}

\bibitem{Garcia2017}
Garcia, D., Lugo, A.E., Hemberg, E., O'Reilly, U.M.: Investigating
  coevolutionary archive based genetic algorithms on cyber defense networks.
  In: Genetic and Evolutionary Computation Conference Companion. pp.
  1455--1462. GECCO (2017)

\bibitem{Hillis1990}
Hillis, W.D.: Co-evolving parasites improve simulated evolution as an
  optimization procedure. Phys. D  42(1--3),  228--234 (Jun 1990),
  \url{https://doi.org/10.1016/0167-2789(90)90076-2}

\bibitem{Miguel2018}
Miguel~Antonio, L., Coello~Coello, C.A.: Coevolutionary multiobjective
  evolutionary algorithms: Survey of the state-of-the-art. IEEE Transactions on
  Evolutionary Computation  22(6),  851--865 (2018)

\bibitem{Pagie2002}
Pagie, L., Mitchel, M.: A comparison of evolutionary and coevolutionary search.
  International Journal of Computational Intelligence and Applications  2(1),
  53--59 (2002)

\bibitem{Popovici2012}
Popovici, E., Bucci, A., Wiegand, R.P., De~Jong, E.D.: Coevolutionary
  principles. In: Handbook of Natural Computing, pp. 987--1033. Springer Berlin
  Heidelberg, Berlin, Heidelberg (2012)

\bibitem{England2016}
{Public Health England}: Government dietary recommendations (August 2016 2016),
  \url{https://www.gov.uk/government/publications/the-eatwell-guide}, pHE
  publications gateway number: 2016202, August 2016. Online [Available]

\bibitem{Rosin1997b}
Rosin, C.D., Belew, R.K.: New methods for competitive coevolution. Evolutionary
  Computation  5(1),  1--29 (1997),
  \url{https://doi.org/10.1162/evco.1997.5.1.1}

\bibitem{Rosin1997}
Rosin, C.D.: Coevolutionary Search Among Adversaries. Ph.D. thesis, Department
  of Computer Science, University of California, San Diego, California (1997)

\bibitem{Simione2021}
Simione, L., Nolfi, S.: Long-term progress and behavior complexification in
  competitive coevolution. Artificial Life  26(4),  409--430 (2021),
  \url{https://doi.org/10.1162/artl_a_00329}

\bibitem{Watson2001}
Watson, R.A., Pollack, J.B.: Coevolutionary dynamics in a minimal substrate.
  In: Genetic and Evolutionary Computation Conference. pp. 702--709. GECCO
  (2001), \url{https://dl.acm.org/doi/10.5555/2955239.2955343}

\bibitem{Wiegand2004}
Wiegand, R.P., Sarma, J.: Spatial embedding and loss of gradient in cooperative
  coevolutionary algorithms. In: Parallel Problem Solving from Nature. pp.
  912--921. PPSN (2004)

\bibitem{WilliamsC2005}
Williams, N., Mitchell, M.: Investigating the success of spatial coevolution.
  In: Genetic and Evolutionary Computation Conference. pp. 523--530. GECCO
  (2005)

\end{thebibliography}

\end{document}